\pdfoutput=1

\documentclass[11pt]{article}

\usepackage{ACL2023}
\usepackage{graphicx}
\usepackage{caption}
\usepackage{subcaption}

\usepackage{times}
\usepackage{latexsym}

\usepackage[T1]{fontenc}

\usepackage[utf8]{inputenc}

\usepackage{microtype}

\usepackage{inconsolata}

\usepackage{listings}
\usepackage{xcolor}

\definecolor{codegreen}{rgb}{0,0.6,0}
\definecolor{codegray}{rgb}{0.5,0.5,0.5}
\definecolor{codepurple}{rgb}{0.58,0,0.82}
\definecolor{backcolour}{rgb}{0.95,0.95,0.92}

\lstdefinestyle{mystyle}{
  backgroundcolor=\color{backcolour},   commentstyle=\color{codegreen},
  keywordstyle=\color{magenta},
  numberstyle=\tiny\color{codegray},
  stringstyle=\color{codepurple},
  basicstyle=\ttfamily\footnotesize,
  breakatwhitespace=false,         
  breaklines=true,                 
  captionpos=b,                    
  keepspaces=true,                 
  numbers=left,                    
  numbersep=5pt,                  
  showspaces=false,                
  showstringspaces=false,
  showtabs=false,                  
  tabsize=2,
  xleftmargin=2em
}

\title{Zshot: An Open-source Framework for Zero-Shot Named Entity Recognition and Relation Extraction}

\author{Gabriele Picco \\
  IBM Research Europe \\
  \texttt{ \small gabriele.picco@ibm.com} \\ \And
  Marcos Martinez Galindo\\
  IBM Research Europe \\
  \texttt{\small marcos.martinez.galindo@ibm.com} \\ \And
    Alberto Purpura\\
  IBM Research Europe \\
  \texttt{ \small alp@ibm.com} \\ \AND 
    Leopold Fuchs\\
  IBM Research Europe \\ 
  \texttt{\small leopold.fuchs@ibm.com} \\ \And 
     Vanessa Lopez\\
  IBM Research Europe  \\
  \texttt{\small vanlopez@ie.ibm.com} \\ \And
   Hoang Thanh Lam\\
  IBM Research Europe  \\
  \texttt{\small t.l.hoang@ie.ibm.com} 
  } 

\begin{document}
\maketitle
\begin{abstract}

The Zero-Shot Learning (ZSL) task pertains to the identification of entities or relations in texts that were not seen during training. ZSL has emerged as a critical research area due to the scarcity of labeled data in specific domains, and its applications have grown significantly in recent years. With the advent of large pretrained language models, several novel methods have been proposed, resulting in substantial improvements in ZSL performance. There is a growing demand, both in the research community and industry, for a comprehensive ZSL framework that facilitates the development and accessibility of the latest methods and pretrained models.
In this study, we propose a novel ZSL framework called Zshot that aims to address the aforementioned challenges. Our primary objective is to provide a platform that allows researchers to compare different state-of-the-art ZSL methods with standard benchmark datasets. Additionally, we have designed our framework to support the industry with readily available APIs for production under the standard SpaCy NLP pipeline. Our API is extendible and evaluable, moreover, we include numerous enhancements such as boosting the accuracy with pipeline ensembling and visualization utilities available as a SpaCy extension.
\url{https://youtu.be/Mhc1zJXKEJQ}

\end{abstract}

\section{Introduction}

Zero-Shot Learning (ZSL) is a machine learning field focused on the study of models able to classify objects or perform tasks that they have not experienced during training. This is achieved by leveraging additional information about the output classes, such as their attributes or descriptions.

ZSL has a wide range of potential applications, since it allows a model to generalize and adapt to new situations without requiring retraining or large amounts of labeled data. This can be particularly useful in real world applications where new classes or categories may be constantly emerging and it would be infeasible to retrain the model every time. ZSL can also be used to classify and predict rare or minority classes that may not have a significant amount of labeled data available for training. 

\begin{figure*}[htb!]
\center{\includegraphics[width=1.0\textwidth]
{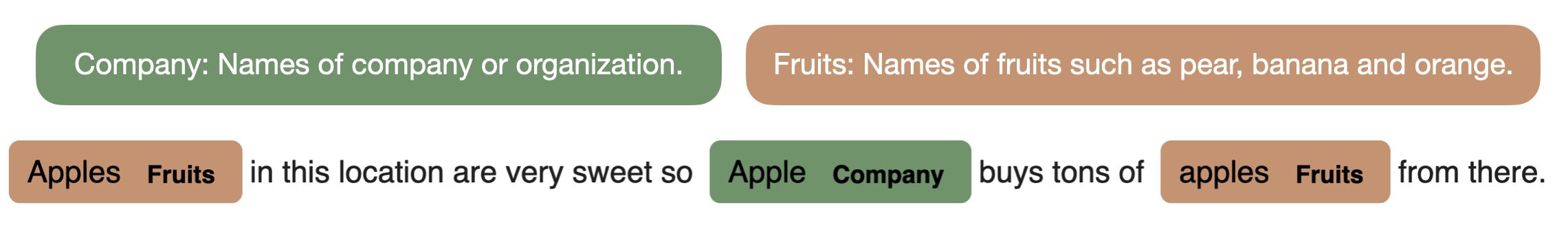}}
\caption{\label{fig:example} An example of a zero-shot NER prediction. Using textual descriptions, the model is able to differentiate between the two unseen classes.}
\end{figure*}

If we consider Named Entity Recognition (NER) -- including classification and linking (NEL) -- and Relation Extraction (RE) problems, recent ZSL methods \citet{aly-etal-2021-leveraging, wu2019zero, DBLP:journals/corr/zs-bert} leverage textual descriptions of entities or relations as additional information to perform their tasks. This additional input allows  models to recognize previously unseen entities (or relations) in text, based on the provided descriptions. \citet{wu2019zero} and \citet{aly-etal-2021-leveraging} provide examples of the effectiveness of descriptions in the zero-shot NER task. The same mechanism can also be applied to the RE task \cite{DBLP:journals/corr/zs-bert} by providing descriptions of the relation between entity pairs.
Figure \ref{fig:example} shows the input and output of a zero-shot NER and classification model such as SMXM \cite{aly-etal-2021-leveraging}. Leveraging entity descriptions, the model is able to disambiguate between mentions of the term \textit{"apple"} -- which could indicate a mention of the homonymous company, or the fruit -- despite having been trained only on OntoNotes \cite{Weischedel2017OntoNotesA} classes. By projecting each token in the latent space of a pre-trained Language Model (LM), the semantic distance between the label/description of a class and each token in a sentence can be used as a method assign tokens to a certain entity. 

There are several variations to ZSL approaches, both for NER and RE. For example, \citet{decao2021autoregressive} show how it is possible to frame the zero-shot NER task as an end-to-end autoregressive generation problem that exploits the probability distribution learned by the LM, while \citet{wu2019zero} uses a combination of a bi-encoder and a cross-encoder for entity linking.
In addition to the different configurations of NER and RE models -- i.e. model architectures, training strategies and datasets -- these components often depend on other NLP tools to process texts and on each other, e.g. RE models often are not end-to-end and require entity spans in a text as part of their input. 

\section{Motivations and Contributions}
The hidden complexity in running NER or RE experiments in the ZSL context and the lack of a shared framework and API to describe the inputs and outputs of such models hampers the reproducibility of many approaches, as well as the ability of the research community to evaluate new models against strong baselines under the same experimental conditions. More specifically, we identified the following major shortcomings for what concerns the ability to evaluate existing NER and RE approaches under the same experimental conditions.

\paragraph{Different assumptions about the task.} Some NER and Entity Linking methods assume that the mentions to link to are given, while others are end-to-end. Similar considerations apply also to RE, where some approaches tackle the task as an end-to-end problem and others as a classification one. Furthermore, some approaches may be limited in terms of performance or scalability and might therefore provide an evaluation only on a restricted subset of a dataset or on a simplified version of their task.
For example, entity linking models are often unable to link entities to a dataset the size of Wikipedia or Wikidata in the same way as many RE approaches cannot scale to hundreds of relation classes.
These differences in the input-output configuration of NER and RE models impact the ability of researchers to evaluate new approaches against existing ones, either for the lack of clarity when describing their evaluation strategy or for the investment required to make the appropriate changes to the models configurations, in order to make them comparable under the same experimental conditions.


\paragraph{Lack of standard datasets.} As the performance of ZSL approaches increases, better and more specific datasets are developed to help researchers evaluate increasingly challenging problems. The release of the FewRel and FewRel 2.0 datasets is an example of this refinement process \cite{gao-etal-2019-fewrel}. In the first version of FewRel, a RE model was expected to assign to each entity pair in a sentence a relation from a list. In other words, there was always a correct class for the model to pick to describe the relation between an entity pair. However, the most frequently occurring scenario in the real world is when two entities are not related or when the correct relation between the entities is missing among the options given by the model. This prompted the release of an updated version of the dataset, which includes the \textit{no relation} class as an option for RE models to choose from. Similar considerations can be made for the evaluation of NER models and domain-specific ones. Another aspect which is often overlooked in many research papers is the lack of a shared implementation of a training, validation and test split for a dataset. This is especially important in the ZSL setting where entities or relation classes should not overlap between training/validation and test dataset splits. Often, zero-shot datasets are obtained from a random split of existing datasets, which were not originally designed to be used for the evaluation of zero-shot approaches and contain overlapping output classes in the respective training/validation and test set. In the process of transforming these datasets the exact split of classes is often not reported and this hampers the reproducibility of the evaluation results presented in a research work.

\paragraph{Different evaluation metrics.} To compare a group of models under the same experimental conditions, different researchers will have to agree on a set of evaluation metrics to employ. This often happens when a new NLP task is introduced, together with a reference dataset and baselines. For NER and RE, the reference metric is often the Macro F1 Score, but there might be models which have a better precision/recall balance depending on the task or that have been evaluated on different, more domain-specific metrics when first described to the research community. These differences make it harder for researchers to benchmark models across different datasets or domains and to evaluate the actual improvements of newly proposed approaches.

To tackle the above-described problems concerning the evaluation and benchmarking of zero-shot NER and RE models, we present Zshot. Zshot is an open-source, easy-to-use, and extensible framework for ZSL in NLP. 

The main contributions of the Zshot framework are the following:
\begin{itemize}
    \item Standardization and modularization: Zshot standardizes and modularizes the NER and RE tasks providing an easy to use and customize API for these models.
    \item Unification of NER and RE: these tasks are often considered separately in existing literature, however this is not the case in real world scenarios where the models are strongly interconnected. In Zshot, users can define unified pipelines for both tasks.
    \item Compatible with SpaCy \cite{spacy2} and the HuggingFace library \cite{wolf2019huggingface}: users define pipelines following SpaCy style NLP pipelines and models are hosted by HuggingFace. 
    \item Evaluation: provides an easy to use and extend evaluation framework for different models and pipelines on standard benchmark datasets for NER and RE.
    \item Visualization: Zshot builds on displaCy capabilities as a visualization tool to display results of NER annotations and RE models.
    \item Ensemble pipeline: Zshot provides simple API for ensembling of NER or RE pipelines using different entity or relation descriptions or models, yielding more accurate results than standalone systems.
    \item Open source: the open source community can customise and extend Zshot adding new models, evaluation metrics and datasets.
    
\end{itemize}

\begin{minipage}{.45\textwidth}
\begin{lstlisting}[language=Python, caption=Python example of a Zshot pipeline configuration., label=lst:PipelineConfig, postbreak=\mbox{\quad \quad \quad \quad \quad \quad}]
nlp_config = PipelineConfig(
    entities=[
        Entity(
            name="Company", 
            description="Names of company or organisation"
        ),
        Entity(
            name="Fruits",
            description="Names of fruits such as pear, banana and orange"
        ),
    ],
    linker=LinkerSMXM(),
)
\end{lstlisting}
\end{minipage}\hfill

The Zshot library implements a pipeline defined by 3 components i.e., \textsl{Mention Detection}, \textsl{Entity Linking} and \textsl{Relation Extraction} and is compatible with SpaCy \cite{spacy2}. It extends the popular displaCy tool for zero-shot entity and relation visualization, and defines an evaluation pipeline that validates the performance of all components, making it compatible with the popular HuggingFace library \cite{wolf2019huggingface}. Zshot aims to provide an easy-to-use solution that can be integrated in production pipelines, while at the same time providing researchers with a framework to easily contribute, validate and compare new state-of-the-art approaches.
We open source all the code, models, metrics and datasets used.~\footnote{Zshot: \url{https://github.com/IBM/zshot}}

\section{Design and Implementation}

\begin{figure*}[thb]
\center{\includegraphics[width=1.0\textwidth]
{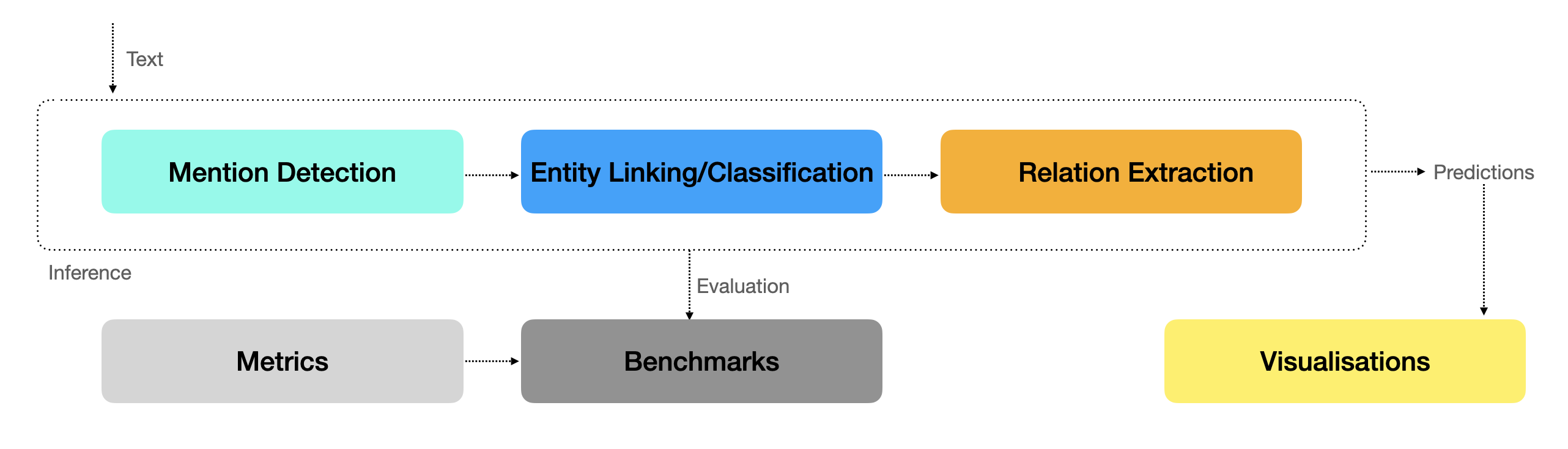}}
\caption{\label{fig:pipeline} High-level architecture of Zshot. The Entity Linker and the Relation Extractor can use the predictions of previous components. For end-to-end models, the pipeline will skip the previous steps.}
\end{figure*}

Figure \ref{fig:pipeline} shows a high-level overview of the Zshot pipeline composed by three main modules i.e., \textsl{Mention Detection}, \textsl{Entity Linking} and \textsl{Relation Extraction}. 
The pipeline accepts a configuration object that allows to select the output classes of mentions, entities and relationships of interest and the models to use to perform each task (see Listing \ref{lst:PipelineConfig}). Then, each of the library modules adds annotations that can be used at a later stage, extending a SpaCy NLP pipeline. Components like the Entity Linker and the Relation Extractor can also be end-to-end, in which case the pipeline automatically skips the unnecessary previous steps. Zshot also automatically manages batching, parallelization and the device on which to perform the computation. All options are configurable and modules are customizable.

Listing \ref{lst:PipelineConfig} contains a sample pipeline configuration for NER, the same used to produce the example shown in Figure \ref{fig:example}.
The configuration defines the two entity classes to be identified, providing a title and a textual description for each of them. It is also specified to use the SMXM end-to-end linker (see Section \ref{ss:entity-linking} for more details).

In the remainder of this section, we will dive deeper into the details of each of the building blocks of Zshot.

\subsection{Mention Detection}

Mention detection is the task that consists in identifying spans of text that may contain entities in a defined set of classes of interest. In our framework, mention detection is used as a pre-processing step for entity linkers (see \ref{ss:entity-linking}) that require mentions as input, such as \cite{wu2019zero} and \cite{cao-2021-holistic}.

Zshot currently supports six mentions extractors. Four reuse existing generic annotators (i.e., Flair \cite{akbik2019flair} and SpaCy) while the other two are sensitive to the defined mentions/entities classes (i.e., TARS \cite{halder-etal-2020-task} and SMXMs).

\subsection{Entity Linking and Classification}
\label{ss:entity-linking}

Entity classification is the process of identifying the type of a given mention in the text. For example, if a text mentions "Barack Obama", entity classification would determine that this refers to a person, but would not specify which person. Entity linking, also known as named entity disambiguation, is the process of identifying and disambiguating mentions of entities in a text, linking them to their corresponding entries in a knowledge base or a dictionary. For example, given "Barack Obama", entity linking would determine that this refers to the specific person with that name (one of the presidents of the United States) and not any other person or concept with the same name.
In Zshot we introduce the Linkers. Linkers can perform both entity classification and entity linking, depending on the entities being used. For simplicity, we use entity linking as the name of the task, but this comprises both entity classification and entity linking. 
Entity linking can be useful for a variety of natural language processing tasks, such as information extraction, question answering, and text summarization. It helps to provide context and background information about the entities mentioned in the text, which can facilitate a deeper understanding of the content.
There are several techniques that can be used for entity linking, including dictionary-based methods, rule-based methods, and machine learning-based methods.

Zshot currently supports four Entity linking and classification methods (Linkers): Blink \cite{wu2019zero}, GENRE \cite{decao2021autoregressive}, TARS and SMXM \cite{aly-etal-2021-leveraging}.

\subsection{Wikification}

Two of the entity linkers, Blink and GENRE, can scale to very large knowledge bases and in Zshot they can be used to perform \textit{Wikification} out-of-the-box.
Wikification is the process of adding \textit{wikilinks} to a piece of text, which allows readers to easily navigate to related articles or pages within a wiki or other online encyclopedia. A wikilink is a hyperlink that is used within Wikipedia -- or another online encyclopedia -- to link to other pages within the same resource.
Wikification is often used to provide context and background information about the concepts and entities mentioned in a piece of text. It can also be used to create a network of interconnected articles within Wikipedia or another online encyclopedia, which makes it easier for readers to explore related topics and gain a deeper understanding of the content.

\subsection{Relation Extraction}

Relaction extraction (or classification) is the task inferring the relation between an entity pair within a portion of text. This task is often framed as a classification problem where a text with entity mentions are given in input to a classifier which is trained to recognize the correct relation type connecting the entities. In the literature, we can also find variants of this approach such as \cite{ni2022generative} which operate in an end-to-end fashion, without requiring entity mentions. ZSL approaches for RE often receive in input a set of candidate of relation descriptions and match them with each entity pair in a text.

Currently, Zshot supports one relation classification model i.e., ZS-BERT~\cite{DBLP:journals/corr/zs-bert}. This model performs relation classification by first embedding relation descriptions using a sentence embedding model and then comparing it with an entity pair embedding computed with a LM fine-tuned for this task. Entity pairs are finally associated to the closest relation class in the embedding space in terms of cosine similarity.

\subsection{Ensembling}
There are different pretrained linkers and mention extractors. These models are pretrained with different data, using different training methods and neural architectures. We provide an easy way to combine the predictions from these models so that we can achieve a more robust result by leveraging the strengths of the diverse set of models. 

Besides the ensembling of linkers, we also support an API for making an ensemble of pipelines with different entity/relation descriptions. We discovered that the accuracy of ZSL models is very sensitive to provided entity/relation descriptions. Combining prediction from pipelines with different descriptions potentially provides significant improvement. An example of the API to make an ensemble of linkers, and mention extractors with various descriptions is demonstrated in the listing \ref{lst:ensemble} in the Appendix.

\subsection{Customization}
One of the most important parts of a framework is to allow the community to create and share new components, which results in improving the models' performance and the appearance of new ones. Zshot allows users to create new components easily, by  extending one of the abstract classes (depending on the type of component: mentions extractor, linker or relation extractor). The user just has to implement the predict method to create a new component, and Zshot will take care of the rest, including adding the result to the \emph{SpaCy} document or adding it to the pipeline. Listing \ref{lst:custom_component} shows a simple example of how to implement a custom mentions extractor that will extract as mentions all the words that contain the letter \emph{s}. The predict method takes as input a list of \emph{SpaCy} documents and return, for each document, a list of Zshot Span, with the boundaries of the mention in the text. We encourage the users to develop new components, models, and datasets and share them with the community. 

\section{Visualization}
\begin{figure}[htb!]
    \centering
         \includegraphics[width=0.5\textwidth]{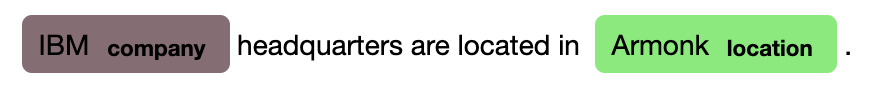}
         \caption{Visualization using Zshot version of displaCy.}
         \label{fig:vis_ner_zshot}
\end{figure}

NER and RE models are broadly used in multiple and diverse NLP pipelines, from Knowledge Graph generation and population to Information Extraction systems. However, for development or research, NER and RE models are sometimes the final output of the process. For these reasons, an appealing visualization of the entities and relations extracted by a model is important. Visualization helps users to rapidly assess what entities and relations have been extracted, and allows them make changes to improve their models. In Zshot, we extend displaCy, a SpaCy tool for visualization. 
When using displaCy for NER with custom entities, the visualization shows the entities, but all of them have the same color, so it's not easy for the user to see the entities detected. We expanded the capabilities of the library to support distinct colors for different entities (see Figure \ref{fig:vis_ner_zshot}). 

\begin{figure}[hbt!]
    \centering
    \includegraphics[width=0.5\textwidth]{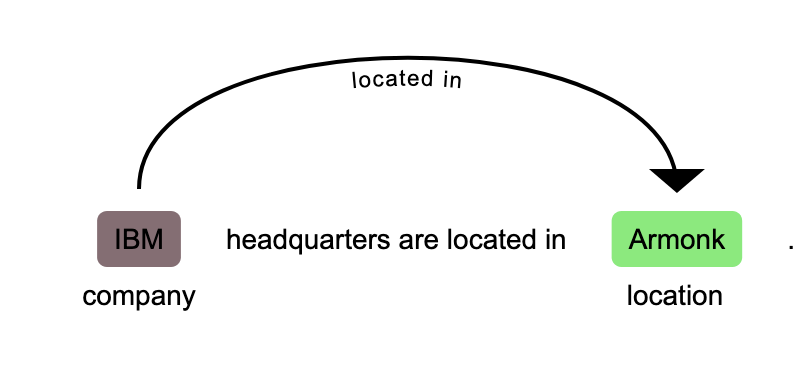}
    \caption{\label{fig:vis_rel_zshot} An example of RE visualization in Zshot.}
\end{figure}

At the time of writing, we could not find a visualization tool for RE in displaCy,~\footnote{\url{https://spacy.io/usage/visualizers}} as RE is not supported in SpaCy. 
In Zshot, we extend displaCy to support the visualization of edges between entity pairs in a sentence as shown in Figure \ref{fig:vis_rel_zshot}.




\section{Evaluation}

Evaluation is key to improve the performance of a system. To assess the performance of NER and RE models, Zshot provides an evaluation module. This module includes an interface to different datasets i.e., OntoNotesZS \cite{pradhan-etal-2013-towards}, MedMentionsZS \cite{mohan2019medmentions} for NER and FewRel \cite{han-etal-2018-fewrel} for RE. All the datasets included have been preprocessed to assure that they can be used in the ZSL setting i.e., the entities/relations in the training or validation splits of these datasets are not overlapping with each other and are not included in the respective test sets to ensure a correct evaluation. These datasets are managed using the Huggingface datasets library,~\footnote{\url{https://github.com/huggingface/datasets}} which makes it easier to add new datasets using the Huggingface Hub~\footnote{\url{https://huggingface.co/datasets}}. 
In Zshot, we use the \texttt{evaluate} package~\footnote{\url{https://github.com/huggingface/evaluate}}, implementing evaluators for the Mentions Extraction, Entity Linking and RE tasks. On top of the evaluators, we added a function that receives a SpaCy NLP model with a Zshot configuration and the splits of the predefined datasets, and it evaluates the model on all datasets configurations. This function returns the evaluation results both as a table with the results as a string object or as a Python dictionary. As a default option, we use the SeqEval package~\cite{seqeval} to compute the evaluation metrics, including Accuracy, Precision, Recall, F1-Score Macro and F1-Score Micro.  However, using the \texttt{evaluate} package, it is also easy to define and use custom metrics extending the same package. Table \ref{tab:metrics} shows an example of the evaluation results format for the OntoNotes validation set, as it is returned by Zshot.

\begin{table}[tbh]
\begin{tabular}{lr}
\hline
{Metric} & {ontonotes-test}\\
\hline
overall\_precision\_macro             & 20.96\%                                   \\
overall\_recall\_macro                & 48.15\%                                  \\
overall\_f1\_macro                    & 29.12\%     
 \\
\hline
\end{tabular}
\caption{\label{tab:metrics} Example of result in Zshot for SMXM linker over the validation set of OntoNotesZS. These are only some metrics reported for one model, to see the whole result please check Table \ref{tab:full-metrics} in the Appendix, with a comparison between two different linkers.}
\end{table}

\section{Conclusion and Future Work}
In this paper, we described Zshot, a framework for zero-shot NER and RE in NLP. ZSL is a growing area of research in NLP. The increasing number of research works published every year and the difficulties associated to the lack of standardization motivated us to develop this framework.

Our work aims at standardizing experimental zero-shot pipelines to improve the reproducibility of existing approaches and facilitate the comparison with new ones. We defined a customizable interface based on the popular SpaCy library. This allows developers and researchers already familiar with this popular Python library to quickly try out zero-shot NLP approaches, while more experienced users can extend it to include new models. We also provide an evaluation package to aid the evaluation and comparison of NER and RE models on different datasets through standard evaluation metrics, which can be further customized and expanded by users.  Finally, we extended the displaCy library to support visualizations of recognized entities and the relations between them as edges between entity spans. We open source all our code, models, metrics and datasets. 

In the future, we plan to increase the number of supported models and datasets for NER and RE. We also aim to increase the efficiency of these models so that they could be employed by NLP practitioners in real world applications.


\section*{Limitations}
Zshot is a SpaCy extension for ZSL. It currently supports models for Mention Detection, Entity Linking and Relation Extraction allowing users to evaluate the supported models and to visualize their outputs. The limitations of our framework are strongly dependent on these models trained on limited amounts of data (in English) for research purposes. Therefore, their results might not be reliable on certain domain-specific scenarios which were not included in the training data and may contain biases. Some models are also more scalable and efficient than others. The efficiency of a model will depend on its implementation. We focus on providing a standard an API and an efficient framework based on SpaCy to run these models. The pretrained models are required to fine-tuned with in-domain training classes, even non-overlapping with testing classes, generalization of these models in new domains is considered as future work.


\section*{Ethics Statement}

Zshot is an opensource, easy-to-use, and extensible framework for ZSL. 
These models may contain bias and cause ethical concerns, as it can be seen in Figure \ref{fig:bias} (see Appendix), where one of the models supported in Zshot assigns an entity label in a biased way, being based on the gender of the name. Bias in NLP models is a common issue \cite{Stanczak2021}, and a lot of efforts focus on mitigating this problem, \cite{sun-etal-2019-mitigating}. Zshot is a framework that aims at standardizing and facilitating the use zero-shot NLP models, and it does not increase nor decrease their bias. We encourage users to assess the bias of NLP models prior to employing them in any downstream task and to validate their results depending on the application scenario to eliminate any potentially negative impact.

\clearpage
\newpage

\bibliography{anthology,custom}
\bibliographystyle{acl_natbib}

\appendix

\newpage
\onecolumn
\section{Appendix: Code Examples}
\label{sec:appendix}

\begin{figure*}[!htb]
    \centering
    \includegraphics[width=1\textwidth]{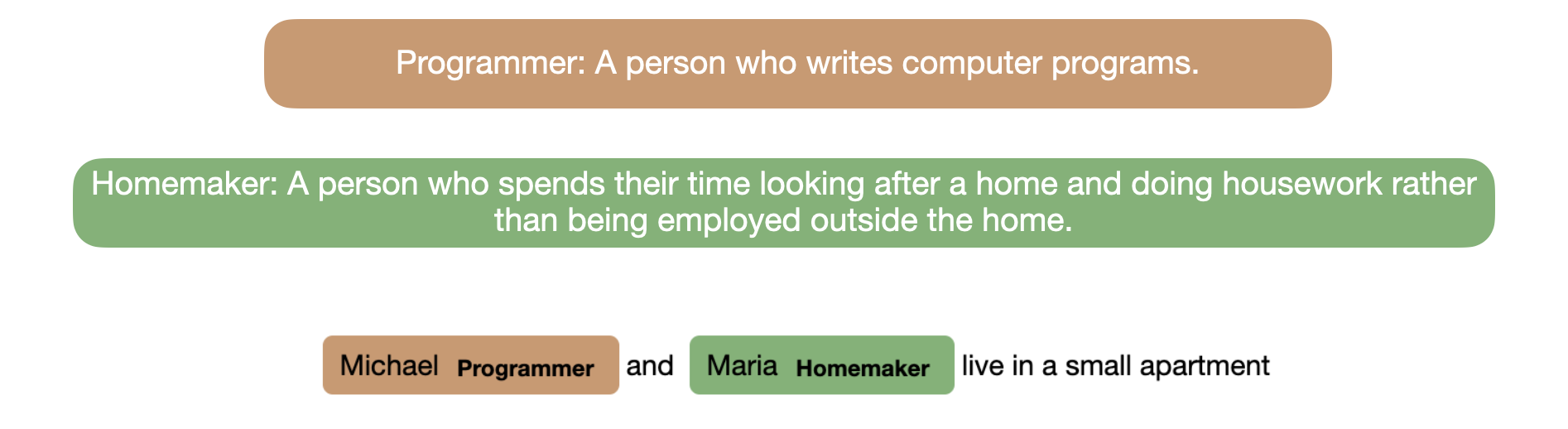}
    \caption{\label{fig:bias} Example of bias of the SMXM linker model integrated in Zshot.}
\end{figure*}

\begin{figure*}[!htb]
    \centering
    \includegraphics[width=1\textwidth]{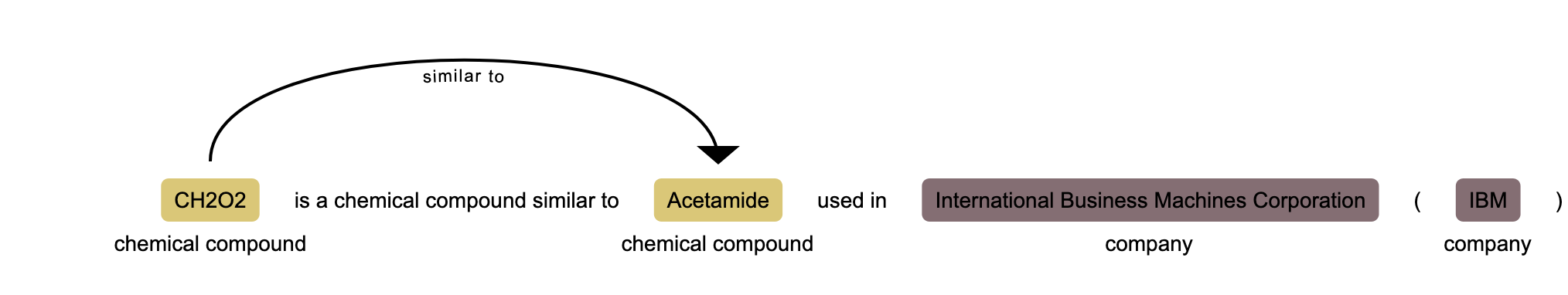}
    \caption{\label{fig:similar} Example of sentence annotation with Zshot. Code of Listing \ref{lst:acetamide} was used to generate this example.}
\end{figure*}

We report below some examples of use of the Zshot framework:

\begin{itemize}
  \item Listing \ref{lst:visualization} shows an example of how to use Zshot to compute both entities and relations and to visualize RE, as shown in Figure \ref{fig:vis_rel_zshot}.
  \item Listing \ref{lst:custom_component} shows an example of how to create a custom mentions extractor in Zshot.
  \item Figure \ref{fig:bias} reports an example of possible biases contained in NLP models for relation classification.
  \item Listing \ref{lst:ensemble} shows an example of how to use Zshot to make an ensembles of existing pretrained linkers and descriptions. In this specific example, two linkers are used together with two different descriptions of  the entity \emph{fruits}.
  \item Listing \ref{lst:acetamide} shows an example of how to use Zshot to annotate a sentence with entities and relations. Figure \ref{fig:similar} shows the result.
\end{itemize}

\begin{figure*}[htb!]
\begin{lstlisting}[language=Python, caption=Python example of using Zshot displacy., label=lst:visualization]
# Import SpaCy
import spacy

# Import PipelineConfig for Zshot configuration and displacy for visualization
from zshot import PipelineConfig, displacy
# Import Entity and Relation data models
from zshot.utils.data_models import Entity, Relation
# Import LinkerSMXM for end2end NER
from zshot.linker import LinkerSMXM
# Import ZSRC for RE
from zshot.relation_extractor import RelationsExtractorZSRC

# Create/Load an NLP model
nlp = spacy.load("en_core_web_sm")
# Create a Zshot configuration with SMXM, ZSRC and some entities and relations
nlp_config = PipelineConfig(
    linker=LinkerSMXM(),
    relations_extractor=RelationsExtractorZSRC(thr=0.1),
    entities=[
        Entity(name="company", description="The name of a company"),
        Entity(name="location", description="A physical location"),
    ],
    relations=[
        Relation(name='located in', description="If something like a person, a building, or a company is located in a particular place, like a city, country of any other physical location, it is present or has been built there")
    ]
)
# Add Zshot to SpaCy pipeline
nlp.add_pipe("zshot", config=nlp_config, last=True)

# Run the pipeline over a text
text = "IBM headquarters are located in Armonk."
doc = nlp(text)
# Visualize the result
displacy.render(doc, style='rel')
\end{lstlisting}
\end{figure*}

\begin{figure*}[htb!]
\begin{lstlisting}[language=Python, caption=Example of creating a custom mentions extractor in Zshot., label=lst:custom_component]
# Import SpaCy
import spacy
# Import types needed for typing
from typing import Iterable
from spacy.tokens import Doc
# Import Zshot PipelineConfig
from zshot import PipelineConfig
# Import Zshot Span to save the results
from zshot.utils.data_models import Span
# Import abstract class
from zshot.mentions_extractor import MentionsExtractor

# Extend MentionsExtractor
class SimpleMentionExtractor(MentionsExtractor):
    # Implement predict function
    def predict(self, 
                docs: Iterable[Doc], 
                batch_size=None) -> Iterable[Iterable[Span]]:
        spans = [[Span(tok.idx, tok.idx + len(tok)) 
                  for tok in doc if "s" in tok.text] for doc in docs]
        return spans

# Create spacy pipeline 
nlp = spacy.load("en_core_web_sm")
# Create config with the new custom
config = PipelineConfig(
    mentions_extractor=SimpleMentionExtractor()
)
# Add zshot with custom component to the spacy pipeline
nlp.add_pipe("zshot", config=config, last=True)

text_acetamide = "CH2O2 is a chemical compound similar to Acetamide used in International Business Machines Corporation (IBM)."
# Run the pipeline and print the result
doc = nlp(text_acetamide)
print([doc.text[mention.start:mention.end] for mention in doc._.mentions])
# -> ['is', 'similar', 'used', 'Business', 'Machines']
\end{lstlisting}
\end{figure*}

\begin{figure*}[htb!]
\begin{lstlisting}[language=Python, caption=Example of annotating a sentence in ZShot., label=lst:acetamide]
from zshot.utils.data_models import Entity
from zshot.linker import LinkerSMXM
from zshot import PipelineConfig, displacy
import spacy

nlp = spacy.blank('en')
config = PipelineConfig(
    entities=[
        Entity(name="company", description="The name of a company"),
        Entity(name="location", description="A physical location"),
        Entity(name="chemical compound", description="any substance composed of identical molecules consisting of atoms of two or more chemical elements."),
    ], 
    linker=LinkerSMXM()
)
nlp.add_pipe("zshot", config=config, last=True)

text_acetamide = "CH2O2 is a chemical compound similar to Acetamide used in International Business " \
        "Machines Corporation (IBM)"

doc = nlp(text_acetamide)
displacy.render(doc, style="ent")
\end{lstlisting}
\end{figure*}

\begin{figure*}[htb!]
\begin{lstlisting}[language=Python, caption=Python example of making an ensemble of pipelines with two linkers and two different descriptions of the entity \emph{fruits}., label=lst:ensemble]
# import necessary libraries and define a spacy pipeline
import spacy
from zshot import PipelineConfig
from zshot.linker import LinkerSMXM, LinkerTARS
from zshot.linker.linker_ensemble import LinkerEnsemble
from zshot.utils.data_models import Entity
from zshot import displacy
nlp = spacy.blank("en")

# a list of two different descriptions of the entity "fruits"
enhanced_descriptions = [[Entity(name="fruits", description="The sweet and fleshy product of a tree or other plant.")],
                       [Entity(name="fruits", description="Names of fruits such as banana, oranges")]]

# a list of two different pretrained linkers
linkers = [LinkerSMXM(), LinkerTARS()]

# Define an ensemble linker with the given enhanced descriptions and pretrained linker models. 
# The provided threshold=0.25 means that at least 1 pipeline out of 4 votes for an output span. 
# We have overall 4 pipelines in the ensemble (2 enhanced descriptions times 2 linkers).
ensemble=LinkerEnsemble(enhance_entities=enhanced_descriptions, linkers=linkers, threshold=0.25)

# add the ensemble to the spacy NLP pipeline
nlp.add_pipe("zshot", config=PipelineConfig(linker=ensemble), last=True)

# annotate a piece of text
doc = nlp('Apples or oranges have a lot of vitamin C.')

# Visualize the result
displacy.render(doc)
\end{lstlisting}
\end{figure*}

\begin{table*}[tbh]
\centering
\begin{tabular}{lrr}
\hline
{Metric} & {SMXM ontonotes-test} & {TARS ontonotes-test} \\
\hline
overall\_precision\_micro                    & 22.12\%  & 34.87\%                               \\
overall\_recall\_micro                       & 50.47\% & 48.22\%                               \\
overall\_f1\_micro                     & 30.76\%      & 40.47\%                                 \\
overall\_precision\_macro               & 20.96\%     & 28.31\%                                  \\
overall\_recall\_macro               & 48.15\%   & 33.64\%                                       \\
overall\_f1\_macro                    & 28.12\%        & 28.29\%                               \\
overall\_accuracy                       & 86.36\%     & 90.87\%                                  \\
total\_time\_in\_seconds                     & 1811.5927  & 613.5401                              \\
samples\_per\_second               & 0.2352       & 0.6943                                    \\
latency\_in\_seconds               & 4.2526        & 1.4402                                      \\
\hline
\end{tabular}
\caption{\label{tab:full-metrics} Full result of ZShot for evaluation. This experiment was performed on a MacBook Pro with an Intel-i7, 64Gb of RAM and no GPU. Results and comparison are only for illustration, as models have not been trained over the same data.}
\end{table*}

\end{document}